\documentclass[letterpaper]{article} 
\usepackage[]{aaai25}  
\usepackage{times}  
\usepackage{helvet}  
\usepackage{courier}  
\usepackage[hyphens]{url}  
\usepackage{graphicx} 
\usepackage{natbib}  
\usepackage{caption} 
\usepackage{pifont}
\usepackage{bbding}
\usepackage{algorithm}
\usepackage{algorithmic}
\usepackage{amsmath}
\usepackage{amsfonts}
\usepackage[misc]{ifsym}

\usepackage{multirow}
\usepackage{makecell}

\usepackage{booktabs}
\usepackage{array}
\usepackage{subfigure}

\newcolumntype{C}[1]{>{\centering\arraybackslash}m{#1}}

\title{FMI-TAL: Few-shot Multiple Instances Temporal Action Localization \\
by Probability Distribution Learning and Interval Cluster Refinement}
\author {
    Fengshun Wang\textsuperscript{\rm 1},
    Qiurui Wang\thanks{Corresponding author}\textsuperscript{\rm 1},
    Yuting Wang\textsuperscript{\rm 1}
}
\affiliations {
    \textsuperscript{\rm 1}Capital University of Physical Education and Sports\\
    \{wangfengshun2023,wangqiurui,wangyuting2023\}@cupes.edu.cn
}

\begin{document}

\maketitle

\begin{abstract}
The present few-shot temporal action localization model can't handle the situation where videos contain multiple action instances. So the purpose of this paper is to achieve manifold action instances localization in a lengthy untrimmed query video using limited trimmed support videos. To address this challenging problem effectively, we proposed a novel solution involving a spatial-channel relation transformer with probability learning and cluster refinement. This method can accurately identify the start and end boundaries of actions in the query video, utilizing only a limited number of labeled videos. Our proposed method is adept at capturing both temporal and spatial contexts to effectively classify and precisely locate actions in videos, enabling a more comprehensive utilization of these crucial details. The selective cosine penalization algorithm is designed to suppress temporal boundaries that do not include action scene switches. The probability learning combined with the label generation algorithm alleviates the problem of action duration diversity and enhances the model's ability to handle fuzzy action boundaries. The interval cluster can help us get the final results with multiple instances situations in few-shot temporal action localization. Our model achieves competitive performance through meticulous experimentation utilizing the benchmark datasets ActivityNet1.3 and THUMOS14. Our code is readily available at \textit{https://github.com/ycwfs/FMI-TAL}.
\end{abstract}

\begin{figure}[!ht]
    \centering
    \subfigure[Other methods]{
        \includegraphics[width=\linewidth]{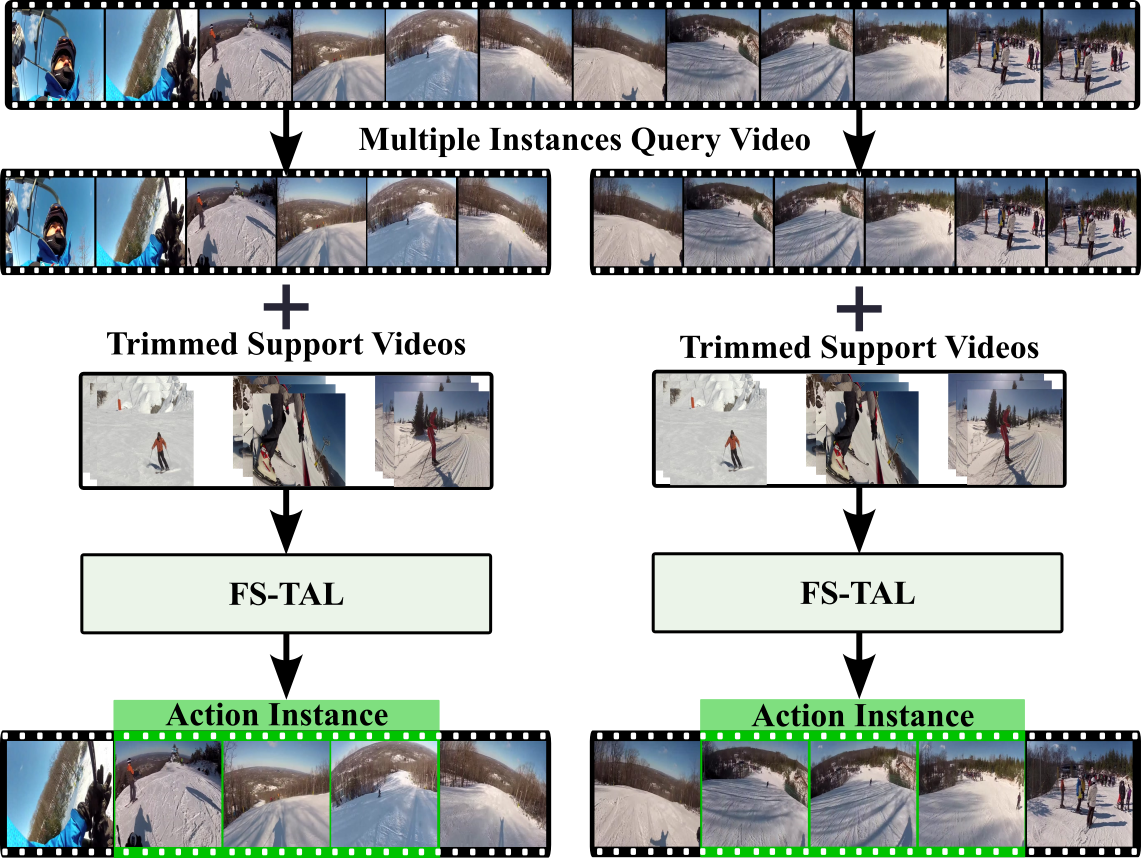}
    }
    \subfigure[Our proposed method]{
        \includegraphics[width=\linewidth]{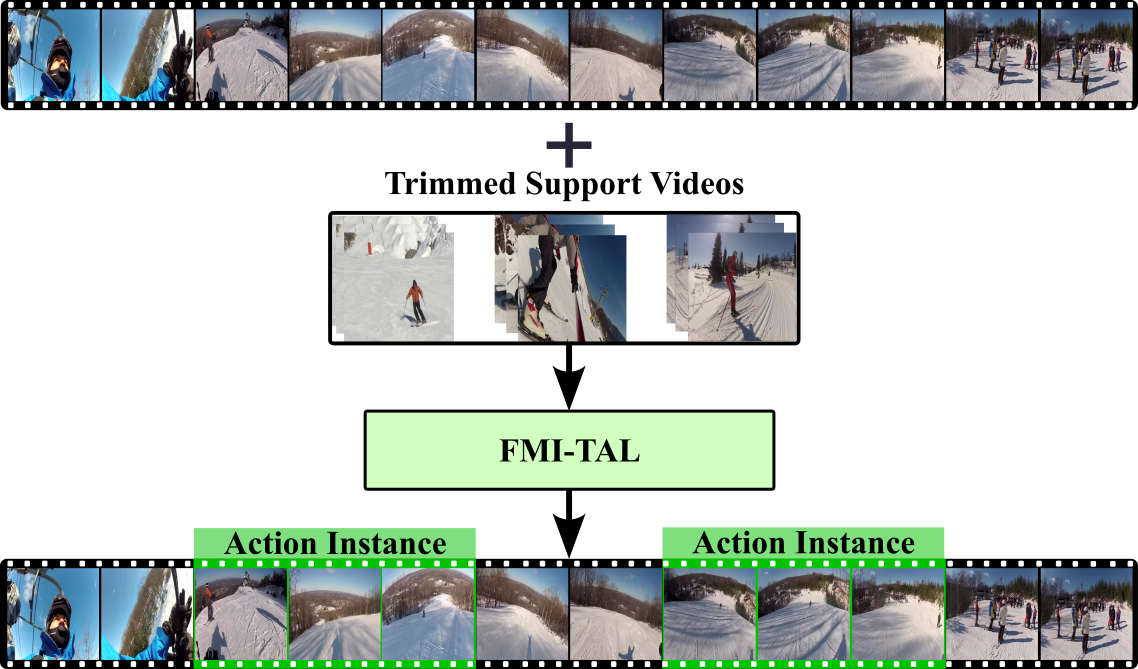}
    }
    \caption{(a) Other methods need to split data first. (b) Our proposed method demonstrates the capability to localize multiple action instances within an untrimmed query video, utilizing a few trimmed support videos. This is achieved without necessitating dataset partitioning.}
    \label{fig:f}
\end{figure}

\section{Introduction}

Few-shot temporal action localization only requires a small number of annotated samples to process and analyze a large amount of unknown video content in real world, which is significant for understanding human behavior, abnormal detection and etc. However, existing few-shot temporal action localization methods achieve the localization of action start time and end time by cutting the video into video segments containing only one action content, which is a practical problem.
  
According to researches \cite{vedaldi_localizing_2020,feng2018video,Yang2018OneShotAL}, Few-Shot Temporal Action Localization (FS-TAL) methods typically use several few videos as support samples for temporal action localization. The purpose of FS-TAL is to identify and locate the same action instances in the given query video. Existing FS-TAL methods aim to alleviate the constraints of time and cost in the annotation of voluminous video datasets, empowering them to swiftly adapt to new classes with only a limited number of additional training videos. 
Although attention or transformer architecture has been used to enhance the performance in recent FS-TAL researches \cite{nag_few-shot_2021,yang_few-shot_2021,Lee_2023_ICCV,Hsieh_2023_WACV}, they still need to split the videos that contain several action instances into several video clips where one video clip has one single action instance. Thus, these existing FS-TAL methods cannot handle a video sample with multiple instances simultaneously in real world. Besides, these approaches directly exploit the extracted features from videos by 3D convolution operations, without considering the relations of extracted features in temporal dimension, spatial dimension and feature dimension.

Unlike previous researches, we propose a real Few-shot Multiple Instances Temporal Action Localization (FMI-TAL) approach. Inspired by \cite{thatipelli2022spatio,perrett2021temporal,yang_few-shot_2021}, the proposed spatial contextual aggregation module and inter-channel dependency module can fully capture the connection among different spatial and channels within each patch region. The encoder and decoder in our method are utilized to learn the temporal relation between query and support videos. In addition, the Selective Cosine Penalization Algorithm is used to restrain improper action instance boundaries. Furthermore, a probability learning process is applied to realize multiple instance temporal action localization learning and prediction after the proposed label distribution generator module converts the original start time and end time of action instance into probability distributions. Finally, the most suitable temporal action boundaries are selected from all the prediction ones based on the prediction probability distributions of action boundaries from the whole network.
The principal contributions can be summarized as follows:
\begin{itemize}
\item We propose a novel Spatial-Channel Relation Transformer (SCR-Transformer) to explore the relations of extracted features in temporal, spatial and channel dimensions, enhancing our method's feature express capability. 
\item A probability learning process is utilized to enable our approach to simultaneously process multiple-instance video without splitting the video into one-instance video clips by hand, enhancing the method's versatility and efficiency in multiple instances of temporal action localization scenarios.
\item Top combinations selection module and Interval cluster module are exploited to acquire the best suitable temporal action boundaries and give state-of-the-art performance compared to the existing FS-TAL methods. 
\end{itemize}

\begin{figure*}[t]
    \centering
    \includegraphics[width=\textwidth]{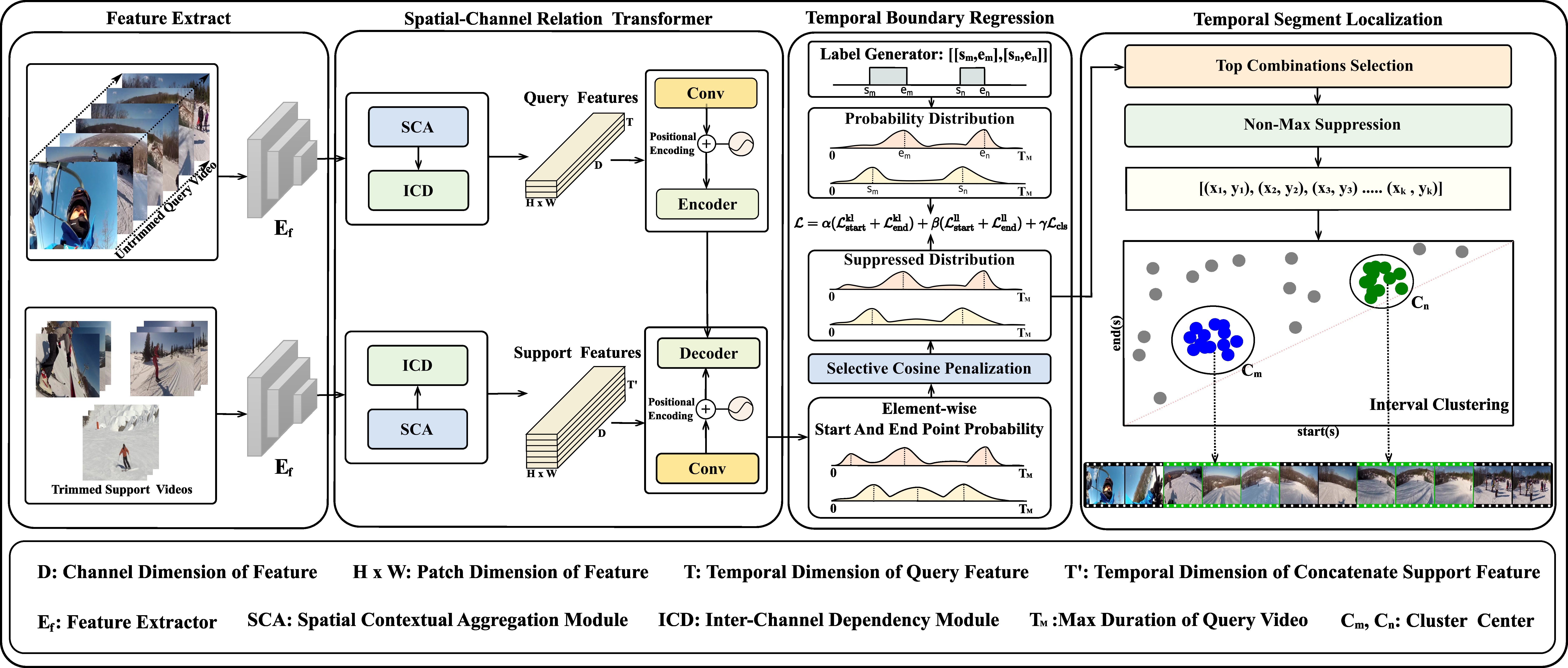}
    \caption{\textbf{Overview of our method}. We first handle and integrate the extracted features by spatial-channel relation transformer. The enhanced features are fed into the Temporal Boundary Regression module to give probability distributions of action boundaries. All probability distributions of action boundaries are selected by the Temporal Segment Localization module to give the best results.}
    \label{fig:overview}
\end{figure*}

\section{Related Works}
\paragraph{Few-shot Temporal Action Localization}
Temporal Action Localization (TAL) aims to precisely locate actions in long and untrimmed videos, playing a crucial role in video comprehension, clip generation, abnormal behavior detection, action quality assessment and etc. The field has evolved from traditional sliding window approaches \cite{shou_temporal_2016,dai2017temporal,gao2018ctap,tran2015learning,chao_rethinking_2018} to more sophisticated methods. Proposal-based method \cite{xu_two-stream_2019,Xu2017RC3DRC} uses regional 3D convolution networks to generate temporal proposals encompassing non-background activity. There are other different mechanisms used to generate proposal \cite{wang_proposal_2021,Liu2018MultiGranularityGF,Yin2023ProposalbasedTA,Tan2021RelaxedTD,Yang2021TemporalAP,Su2023MultiLevelCB}. Based on these proposal-based methods, \cite{lin2018bsn} propose the Boundary Sensitive Network (BSN) to generate high-quality temporal proposals by modeling boundary probability and evaluating proposal confidence. Furthermore, graph convolutional based approaches \cite{zeng_graph_2019,zeng_graph_2022,huang2020improving,tang2023ddg,Gan2023TemporalvisualPG} models build relations between temporal proposals and capture long-range dependencies. There are also some researches \cite{zhang_learning_2020,yang_revisiting_2020,zhao_temporal_2020,Yuan2016TemporalAL,He2021FeaturePH,Li2022PyramidRS,Xia2023ExploringAC} focusing on anchor and feature pyramid mechanism. Recent researches have incorporated attention mechanisms and Transformer architectures \cite{liu_end--end_2022,chen_relation_2020,gao_novel_2023,yin_enhanced_2023,Gan2023TemporalAP,zhang2022actionformer} to capture global contextual information and improve localization accuracy. Despite of above methods, fully-supervised approaches remain limitations to localizing actions since the annotation can hardly be obtained in real world. 
Few-shot learning (FSL) addresses this limitation by learning the inner regulars by using only a small set of samples. It is particularly useful for Temporal Action Localization when considering the huge amount of videos and time consumption of annotations. Foundational work in FSL includes prototypical networks \cite{snell2017prototypical}, which represent classes by prototypes computed as the mean of their examples in a learned representation space. \cite{vinyals2016matching} proposed Matching Networks, utilizing a differentiable nearest neighbor algorithm for few-shot learning. The relation network by \cite{sung2018learning} further extends this idea by learning a deep distance metric to compare query images with few-shot examples. The integration of FSL with TAL is defined as Few-shot Temporal Action Localization (FS-TAL), pioneered by \cite{feng2018video} which proposes locating semantically corresponding segments between query and reference videos using limited examples. FS-TAL combines the temporal precision of TAL with the adaptability of FSL, allowing models to localize actions in videos with minimal annotated data and extend to new, unseen action classes. This approach not only addresses the data scarcity issue in video annotation but also enhances the generalization capabilities of action localization systems, making them more applicable to real-world scenarios where new actions may frequently emerge.

\paragraph{Probability Distribution Learning}
Probability distribution learning estimates the underlying probability distribution of data \cite{baum1987supervised}, enhancing temporal action localization by capturing uncertainty in action boundaries. Nag et al. \cite{nag2023post} propose GAP where the method address temporal quantization errors in TAL caused by video downsampling. GAP models action boundaries with a Gaussian distribution and uses Taylor expansion for efficient inference, improving TAL performance without needing model changes or retraining.


\section{Methodology}

In FS-TAL, we address a dataset $D$ of untrimmed videos, where actions are annotated with start time, end time, and class labels from a space $Y$, partitioned into $Y_{train}$, $Y_{test}$, and $Y_{val}$. Given a support set $S=\{(V_i,t_{start_i},t_{end_i},y_i)|y_i\in Y_{test}\}_{i=1}^N$ and a query video $Q$, our objective is to predict temporal intervals and labels for actions in $Q$, outputting $P=\{(\hat{t}_{start_j},\hat{t}_{end_j},\hat{y}_j)|\hat{y}_j\in Y_{test}\}_{j=1}^M$. Firstly, A 3D convolution network extracts features from both query and support videos. These features are processed by a spatial-channel relation transformer (SCR-Transformer) and a mask convolutional projection. The SCR-Transformer includes spatial contextual aggregation, inter-channel dependency, and feature relation transformation modules. The Selective Cosine Penalization algorithm enhances \textit{softmax} probabilities, and the loss is computed against pre-generated labels. This framework enables effective localization of actions in untrimmed query video based on trimmed support videos.

\subsection{Feature Extractor}
We use a pre-trained C3D \cite{tran2015learning} backbone to extract features for both query and support videos. For the uncut query video, features are represented as $F^q \in \mathbb{R}^{HWC \times T}$, where $H$, $W$, $C$, and $T$ are the height, width, channel, and temporal dimensions, respectively. Each clip feature at index $i$ is denoted as $F_i^q \in \mathbb{R}^{HWC}$. Support features are extracted similarly but concatenated along the temporal dimension: $F^s \in \mathbb{R}^{HWC \times T'}$, with $T' = t_1 + t_2 + ... + t_n$. This provides both query and support features for further processing.

\subsection{Spatial-Channel Relation Transformer}

\subsubsection{Spatial Contextual Aggregation Module}
We utilize the spatial contextual aggregation module to enrich spatial contextual semantics and capture spatial relationships among different patches within each frame. The input query and support feature tensors are defined as $X \in \mathbb{R}^{T \times N \times D}$ and $X' \in \mathbb{R}^{T' \times N \times D}$, where $T$ and $T'$ represent the temporal dimension, $N$ represents the spatial dimension, and $D$ represents the feature channel dimension.

Positional embedding  $\text{PE}(\cdot)$ is first applied to the input features $X$ to incorporate positional information:
\begin{equation}
\tilde{X} = X + \text{PE}(X),
\end{equation}
Next, the embedded features $\tilde{X}$ are passed through three linear projection layers to generate the query (Q), key (K), and value (V) vectors:
\begin{equation}
Q = \text{L}_Q(\tilde{X}), \quad K = \text{L}_K(\tilde{X}), \quad V = \text{L}_V(\tilde{X}),
\end{equation}
The attention score matrix $A \in \mathbb{R}^{T \times N \times N}$ is computed with scaling factor $\sqrt{D}$ by:
\begin{equation}
A = \text{Softmax}\left(\frac{QK^T}{\sqrt{D}}\right),
\end{equation}
Finally, the attention-weighted value is computed by using the attention score matrix $A$ and the value vector $V$, and added to the original input features $X$ to obtain the enhanced spatial features:
\begin{equation}
\text{Output} = \gamma \cdot AV + X,
\end{equation}
where $\gamma$ is a learnable scaling parameter. The output of the spatial attention module, $\text{Query Output} \in \mathbb{R}^{T \times N \times D}$ and $\text{Support Output} \in \mathbb{R}^{T' \times N \times D}$ will be fed into subsequent modules for further processing.

\subsubsection{Inter-Channel Dependency Module}
The inter-channel dependency (ICD) module captures correlations among different channels within each patch region. The input feature tensors are defined as $X^{q} \in \mathbb{R}^{T \times N \times D}$ and $X^{s} \in \mathbb{R}^{T' \times N \times D}$, where $T$ and $T'$ represent the temporal dimension, $N$ the spatial dimension, and $D$ the channel dimension.

The ICD module consists of channel fusion and channel linear sub-modules. The channel fusion module reshapes the input tensor $X$ to $\mathbb{R}^{T \times D \times N}$ and applies a 1D convolution for the channel dimension:
\begin{equation}
Y = (\mathcal{F}_D(X^\top))^\top,
\end{equation}
where $\mathcal{F}_D$ denotes a 1D convolution with a kernel size of 1, acting on the channel dimension $D$.

The channel linear module applies a non-linear transformation to $Y$ using two linear layers with a ReLU activation function in between:
\begin{equation}
A = \varrho_{out}(\text{ReLU}(\varrho_{in}(Y))),
\end{equation}
where $\varrho_{out}$ and $\varrho_{in}$ are linear layers with input and output dimensions equaling to $D$.

Finally, $A$ is added to the original input features $X$ via a residual connection:
\begin{equation}
\text{Output} = X + A.
\end{equation}

The resulting tensor $\text{Output} \in \mathbb{R}^{T \times D \times N}$, integrating channel-wise attention, serves as input for subsequent processing modules.


\subsubsection{Feature Relation Transformation}
After obtaining spatially and channel-related features, we apply 1D convolution to reduce dimensions:
$\boldsymbol{x} = \text{Conv1D}(\text{X}_q)$, $\boldsymbol{y} = \text{Conv1D}(\text{X}_s)$
This yields query sequence $\boldsymbol{x} \in \mathbb{R}^{T \times D}$ and support sequence $\boldsymbol{y} \in \mathbb{R}^{T^{'} \times D}$.
Our Transformer, comprising an encoder and decoder based on the standard Transformer architecture, processes these sequences. The encoder contextualizes the query sequence:
$\boldsymbol{h}^{enc} = \text{Encoder}(\boldsymbol{x})$, where $\boldsymbol{h}^{enc} \in \mathbb{R}^{T \times D}$
The decoder then integrates the encoded representation with the support sequence:
$\boldsymbol{h}^{dec} = \text{Decoder}(\boldsymbol{h}^{enc}, \boldsymbol{y})$, where $\boldsymbol{h}^{dec} \in \mathbb{R}^{T \times D}$. 
Both the encoder and decoder consist of multiple layers with attention mechanisms and feed-forward networks. The final output $\boldsymbol{h}^{dec}$ maintains the temporal dimension $T$ of the query video.

\subsection{Temporal Boundary Regression}
After we get the probability sequence of the SCR-Transformer, Subsequently, we first construct a random tensor $\boldsymbol{V} \in \mathbb{R}^{\text{T}_{max} \times D}$, where $\text{T}_{max}$ denotes the longest duration of video seconds in all datasets. Then we set the value of $\boldsymbol{V}$ to 0 when $idx$ is larger than $T$ to mask the absent time steps and copy the original value of $\boldsymbol{h}^{dec}$ to $\boldsymbol{V}$ when $idx$ is smaller than $T$. Then the tensor $\boldsymbol{V}$ is passed through three separate linear projection modules $\phi$ to generate the start timestamp, end timestamp and classification scores, The start timestamp score $\boldsymbol{s}$ and end timestamp score $\boldsymbol{e}$ are passed through a \textit{softmax} layer to obtain probability distributions over the sequence length. The classification scores $\boldsymbol{c}$ are used for classification tasks:
\begin{equation}
\begin{aligned}
\boldsymbol{S_s} &= \text{Softmax}(\Phi_s(\boldsymbol{V})), \\
\boldsymbol{S_e} &= \text{Softmax}(\Phi_e(\boldsymbol{V})), \\
\boldsymbol{S_c} &= \Phi_c(\boldsymbol{V}),
\end{aligned}
\end{equation}
This allows our model to handle variable length inputs without needing predefined feature pyramids, time intervals etc.

\subsubsection{Selective cosine penalization}
We propose a novel Selective Cosine Penalization (SCP) Algorithm to refine the preliminary probabilities $\in \mathbb{R}^{T}$ and more accurately locate action segments. SCP selectively represses temporal boundaries by leveraging cosine similarities between query features at different time points inside one action instance where the surrounding features for these frames are similar.

The algorithm firstly sorts the start and end probabilities and then calculates cosine similarities between features at specific time intervals. It uses a dynamic threshold based on mean similarities to filter and adjust the probabilities, rather than relying on manually specified values. This approach allows SCP to adapt to different scenes and reduce potential disturbances. 
The detailed process of SCP is presented in Algorithm \ref{scp}, which outlines the step-by-step procedure for probability refinement.

\begin{algorithm} [t]
    \caption{Selective Cosine Penalization Algorithm} \label{scp}
    \begin{algorithmic}[1]
        \REQUIRE
            \parbox[t]{\dimexpr\linewidth-\algorithmicindent}{
                start probabilities $\text{sp} \in \mathbb{R}^{T}$, \\
                end probabilities $\text{ep} \in \mathbb{R}^{T}$, \\
                query features $\text{qf} \in \mathbb{R}^{C \times T \times H \times W}$
            }
        \ENSURE 
        refined start and end probabilities $\text{sp}, \text{ep} \in \mathbb{R}^{T}$
        
        \STATE $\text{acs} \gets [\ ], \text{ace} \gets [\ ]$
        
        \FOR{$\text{idx} \in \text{sp}$}
            \STATE $\text{s} \gets \text{CosineSimilarity}(\text{qf}[:, \text{idx},:,:], \text{qf}[:, \text{idx}-4, :,:])$
            \STATE $\text{acs.append}((\text{idx}, \text{s}))$
        \ENDFOR
        
        \FOR{$\text{idx} \in \text{ep}$}
            \STATE $\text{e} \gets \text{CosineSimilarity}(\text{qf}[:, \text{idx}, :,:], \text{qf}[:, \text{idx}+4, :,:])$
            \STATE $\text{ace.append}((\text{idx}, \text{e}))$
        \ENDFOR
        
        \STATE $\text{mcs} \gets \text{Mean}(\text{acs})$, $\text{mce} \gets \text{Mean}(\text{ace})$
        \STATE $\text{sp[idx]} \gets \text{sp[idx]/2}$ \textbf{if} $\text{acs} < \text{mcs}$
        \STATE $\text{ep[idx]} \gets \text{ep[idx]/2}$ \textbf{if} $\text{ace} < \text{mce}$
        
    \end{algorithmic}
\end{algorithm}

\subsection{Temporal Segment Localization}
We get the final segments by Top Combinations Selection (TCS), Non-max suppression (NMS), and Interval Clustering (IC). This step is crucial for refining the model's predictions and obtaining the most probable temporal segments.
\subsubsection{Top combinations selection}
A score matrix $S \in \mathbb{R}^{T \times T}$ is computed, where each element $S_{ij}$ is the product of the start probability at time $i$ and the end probability at time $j$:
\begin{equation}
    \boldsymbol{S_{ij}} = \boldsymbol{S_s[i]} \cdot \boldsymbol{S_e[j]},
\end{equation}

The algorithm then selects the top-$k$ scores from this matrix. The corresponding indices are converted back to pairs of start and end points $(i,j)$.

To refine these selections, a soft Non-Maximum Suppression (NMS) \cite{neubeck2006efficient} process is applied. This step guarantees that the final set of predictions comprises varied and non-overlapping temporal segment pairs, with the end time occurring after the start time in each pair. The NMS process considers the temporal intersection over union (IOU) \cite{Rezatofighi_2019_CVPR} between segments and eliminates highly overlapping predictions.

The output of NMS is a list of temporal segments, each represented by a start and end point $(i,j)$. These segments represent the most confident and diverse predictions from the model, balancing high probability scores with minimal overlap between segments.
\subsubsection{Interval Clustering}
To further refine our temporal segment prediction and explore alternative approaches, we develop a module Interval Cluster (IC) that treats each predicted time interval as a two-dimensional data point. This module offers a more holistic view of the temporal segments by simultaneously considering start and end times. The method can be described as follows:

\textit{Interval Representation}. Each predicted temporal segment is represented as a two-dimensional point, where the x-coordinate corresponds to the start time and the y-coordinate to the end time. This representation preserves the inherent relationship between start and end times within each prediction.

\textit{Two-Dimensional Clustering}. We employ the DBSCAN (Density-Based Spatial Clustering of Applications with Noise) algorithm \cite{ester1996density} to cluster these two-dimensional points. This clustering step identifies groups of similar temporal predictions in the start-end time-space.

\textit{Cluster Analysis}. For each identified cluster (excluding noise points), we compute the centroid by averaging the start and end times of all intervals within the cluster. This centroid represents the optimal temporal segment for that cluster.

\subsection{Optimization}
\subsubsection{Label generator}
To optimize our model, we design a label generator based on the probability distribution. Firstly, we convert the action segment labels [[$s_1$, $e_1$],[$s_2$, $e_2$] ... [$s_n$, $e_n$]] to two Gaussian Probability Distribution (GPD) called $P_{s} \in \mathbb{R}^{T}$ and $P_{e}\in \mathbb{R}^{T}$ with the parameters of length, center, width, which can be described as Algorithm \ref{gmgpdn}. 

In order to use $P_{s}$ and $P_{e}$ distributions to guide our model learning, we adopt Kullback-Leibler divergence loss $\mathcal{L}_{kl}$ \cite{he2019bounding} and l1 loss $\mathcal{L}_{l1}$. Then for the action classification, we employ focal loss \cite{lin2017focal} as a regularizing mechanism. This technique effectively addresses the class imbalance issue by dynamically adjusting the weights of positive and negative samples. It enables fine-grained control over the contributions of difficult and easy samples to the overall loss, making a improved model performance. Therefore, our overall loss function can be described as: 
\begin{equation}
    \boldsymbol{\mathcal{L}}=\alpha (\boldsymbol{\mathcal{L}^\text{kl}_\text{start}} + \boldsymbol{\mathcal{L}^\text{kl}_\text{end}}) + \beta (\boldsymbol{\mathcal{L}^\text{l1}_\text{start}} + \boldsymbol{\mathcal{L}^\text{l1}_\text{end}}) + \boldsymbol{\gamma\mathcal{L}_\text{cls}},
\end{equation}
The $\alpha$, $\beta$ and $\gamma$ are parameters that are designed to balance the different parts of loss $\boldsymbol{\mathcal{L}}$. Notice that the following conditions should be satisfied: $\alpha + \beta + \gamma = 1$.

\begin{algorithm}[t]
    \caption{Label Generator} \label{gmgpdn}
    \begin{algorithmic}[1]
        \REQUIRE 
            \parbox[t]{\dimexpr\linewidth-\algorithmicindent}{
            Length of sequence $L \in \mathbb{N}$,\\ 
            Labels $S = \{(s_1, e_1), \ldots, (s_i, e_i)\}$, $i \in \mathbb{N}$,\\ 
            Sigma percentage $sp \in \mathbb{R}$,\\
            Noise level $\alpha \in \mathbb{R}$,\\
            Probability threshold for adding noise $\theta \in \mathbb{R}$
            }
        \ENSURE Probability distribution $\mathbf{p} \in \mathbb{R}^L$
        \STATE $\mathbf{p} \gets \mathbf{0}^L$ \COMMENT{Initialize probability distribution}
        \STATE $\mathbf{x} \gets [0, 1, \ldots, L-1]$
        
        \FOR{$(s, e) \in \mathbb{N}$}
            \STATE $w \gets e - s + 1$
            \STATE $\mu \gets (s + e) / 2$
            \STATE $\sigma \gets w * sp$
            \FOR{$i=0$ \TO $L-1$}
                \STATE $\mathbf{p}[i] \gets \mathbf{p}[i] + \exp\left(-\frac{1}{2}\left(\frac{\mathbf{x}[i] - \mu}{\sigma}\right)^2\right)$
            \ENDFOR
        \ENDFOR
        \STATE $\mathbf{p} \gets \text{Smooth}(\mathbf{p})$ \COMMENT{Apply smoothing}
        \STATE $\mathbf{p} \gets \mathbf{p} / (\sum_{i=0}^{L-1} \mathbf{p}[i] + \epsilon)$
        \STATE $\mathbf{n} \gets \text{UniformRandom}(0, \alpha, L)$ \COMMENT{Generate noise}
        \FOR{$i=0$ \TO $L-1$}
            \IF{$\mathbf{p}[i] < \theta$}
                \STATE $\mathbf{p}[i] \gets \mathbf{p}[i] + \mathbf{n}[i]$ 
            \ENDIF
        \ENDFOR
        \STATE $\mathbf{p} \gets \mathbf{p} / (\sum_{i=0}^{L-1} \mathbf{p}[i] + \epsilon)$
        \RETURN $\mathbf{p}$
    \end{algorithmic}
\end{algorithm}

\begin{table*}[t]
\centering

\resizebox{\textwidth}{!}{%
\begin{tabular}{@{}c|c|cccc|cccc@{}}

\toprule

\multirow{3}{*}{Method} & \multirow{3}{*}{Shot} & \multicolumn{4}{c|}{ActivityNet-v1.3} & \multicolumn{4}{c}{THUMOS'14} \\

\cmidrule{3-10}

 &  & \multicolumn{2}{c|}{Single-instance} & \multicolumn{2}{c|}{Multi-instance} & \multicolumn{2}{c|}{Single-instance} & \multicolumn{2}{c}{Multi-instance} \\

\cmidrule{3-10}
 
 &  & mAP@0.5 & mean & mAP@0.5 & mean & mAP@0.5 & mean & mAP@0.5 & mean \\

\midrule

\citeauthor{nag_few-shot_2021} & 1 & 55.6 & 31.8 & 44.9 & 25.9 & 51.2 & 27.0 & 9.1 & 5.3 \\
\citeauthor{Lee_2023_ICCV}& 1 & 62.1 & - & 48.2 & - & 53.8 & - & 9.8 & - \\
\citeauthor{vedaldi_localizing_2020} & 1 & 53.1 & 29.5 & 42.1 & 22.9 & 48.7 & - & - & - \\
\citeauthor{hu2019silco} & 1 & 41.0 & 24.8 & 29.6 & 15.2 & - & - & - & - \\
\citeauthor{feng2018video} & 1 & 43.5 & 25.7 & 31.4 & 17.0 & - & - & - & - \\
\citeauthor{yang_few-shot_2021} & 1 & 57.5 & - & 47.8 & - & - & - & - & - \\
\citeauthor{Hsieh_2023_WACV} & 1 & 60.7 & - & - & - & - & - & - & - \\
\midrule
{\textbf{Ours}}& 1 & \textbf{68.4} & \textbf{37.8} & \textbf{64.2} & \textbf{33.5} & \textbf{58.3} & \textbf{32.4} & \textbf{23.9} & \textbf{11.2} \\

\bottomrule
\end{tabular}%
}
\caption{Results comparison with state-of-the-art under 1-shot learning}
\label{tab:fs-tal-results-1s}
\end{table*}

\begin{table*}[t]
\centering
\resizebox{\textwidth}{!}{%
\begin{tabular}{@{}c|c|cccc|cccc@{}}

\toprule
\multirow{3}{*}{Method} & \multirow{3}{*}{Shot} & \multicolumn{4}{c|}{ActivityNet-v1.3} & \multicolumn{4}{c}{THUMOS'14} \\
\cmidrule{3-10}
 &  & \multicolumn{2}{c|}{Single-instance} & \multicolumn{2}{c|}{Multi-instance} & \multicolumn{2}{c|}{Single-instance} & \multicolumn{2}{c}{Multi-instance} \\
\cmidrule{3-10}
 &  & mAP@0.5 & mean & mAP@0.5 & mean & mAP@0.5 & mean & mAP@0.5 & mean \\
\midrule
\citeauthor{hu2019silco} & 5 & 45.4 & 27.0 & 38.9 & 20.9 & 42.2 & 22.8 & 6.8 & 3.1 \\
\citeauthor{vedaldi_localizing_2020} & 5 & 56.5 & 34.9 & 43.9 & 24.5 & 51.9 & 29.3 & 8.6 & 4.4 \\
\citeauthor{nag_few-shot_2021} & 5 & 63.8 & 38.5 & 51.8 & 30.2 & 56.1 & 32.7 & 13,8 & 7.1 \\
\citeauthor{Lee_2023_ICCV}& 5 & 66.3 & - & 53.5 & - & 59.2 & - & 15.7 & - \\
\citeauthor{yang_few-shot_2021} & 5 & 60.6 & - & 48.7 & - & - & - & - & - \\
\citeauthor{Hsieh_2023_WACV} & 5 & - & - & 61.2 & - & - & - & - & - \\
\midrule

{\textbf{Ours}}& 5 & \textbf{70.2} & \textbf{41.2} & \textbf{67.5} & \textbf{36.6} & \textbf{60.3} & \textbf{36.4} & \textbf{26.8} & \textbf{15.3} \\
\bottomrule
\end{tabular}%
}
\caption{Results comparison with state-of-the-art under 5-shot learning}
\label{tab:fs-tal-results-5s}
\end{table*}

\section{Experiments and Results}

\subsection{Datasets}

We use the benchmarks ActivityNet1.3 \cite{caba2015activitynet} and THUMOS14 \cite{THUMOS14} dataset to evaluate our few-shot action localization model. 

ActivityNet1.3 contains 203 activity classes, averaging 137 untrimmed videos per class and 1.41 activity instances per video in total 849 video hours. It enables comparison of algorithms in uncut video classification, trimmed activity classification, and activity localization.
THUMOS14 is a key benchmark for action localization algorithms. Its training set contains 13,320 videos. The validation, testing, and background sets include 1,010, 1,574, and 2,500 untrimmed videos, respectively. The temporal action localization task covers over 20 hours of video across 20 sports categories. This task is challenging due to the high number of action instances per video and the significant presence of background content (71\% of frames).

Unlike \cite{vedaldi_localizing_2020} and \cite{feng2018video}, we don't remove videos longer than 768 frames in ActivityNet. We also randomly split the classes of the dataset into three subsets at the proportion 7:2:1 for training, validation, and testing.

\begin{table}[t] 
\setlength{\leftskip}{0pt plus 1fil minus \marginparwidth}
\setlength{\rightskip}{\leftskip}
\begin{tabular}{C{2cm}C{2cm}C{2cm}}
\toprule 
\multirow{2}{*}{channel dim} & \multicolumn{1}{c}{1-shot} & \multicolumn{1}{c}{5-shot} \\
\cmidrule{2-3}
 & mAP@0.5 & mAP@0.5 \\
\midrule 
512 & 67.3 & 68.5 \\
2048 & \textbf{68.4} & \textbf{70.2} \\
\bottomrule 
\end{tabular}
\caption{Influence of features' channel dimension}
\label{tab:cd}
\end{table}

\subsection{Implementation Details}
C3D features are extracted by a backbone pre-trained on action recognition using ActivityNet1.3 and THUMOS14 datasets. The input temporal dimension is set to 30 frames, aligning with the video fps. A $256\times{}256$ image input size is used for the C3D network. Data augmentation incorporates random cropping and horizontal mirroring. The video features' patch number is set to $4\times{}4$, without cutting all videos to the same clips as in \cite{nag_few-shot_2021,vedaldi_localizing_2020}.
The model is tested utilizing hydra package \cite{Yadan2019Hydra} for hyperparameters. Training is conducted for 20000 epochs with a batch size of 1, the initial learning rate of $1e-6$, and Adam optimizer (weight decay $5e-4$). A learning rate scheduler is employed, reducing the rate by 0.1 every 5 epochs.
The label generator's sigma is set to 0.1 for THUMOS14 and 0.5 for ActivityNet1.3 and $\epsilon$ is set to $1e-8$. The noise level and probability threshold are fixed at 0.01. Top-k combinations are set to 500, with an NMS threshold of 0.9. DBSCAN parameters (eps, min\_sample) are configured as (3, 2) for THUMOS14 and (5, 2) for ActivityNet1.3.

\subsection{Result Comparison}
For demonstrating our model's effectiveness, we compare it with several state-of-the-art FS-TAL methods, including attention-based \cite{Lee_2023_ICCV,Hsieh_2023_WACV}, transformer-based \cite{yang_few-shot_2021,nag_few-shot_2021}, proposal-based \cite{vedaldi_localizing_2020} models and a few-shot object detection model\cite{Hsieh_2023_WACV}. As shown in Table \ref{tab:fs-tal-results-1s} and Table \ref{tab:fs-tal-results-5s}, our model demonstrates highly competitive performances in the 1-shot and 5-shot scenarios, surpassing all existing methods. In this unified approach, we achieve dominant performance across both single and multi-instance scenarios. 

Our model consistently performs well across various settings, highlighting its efficacy in capturing temporal and spatial information, offering an efficient solution for few-shot temporal action localization tasks in real-world scenarios.

\subsection{Ablation Study}
\subsubsection{The influence of features' channel dimension}
In Table \ref{tab:cd}, we present an ablation study examining the influence of the feature channel dimension on our model's performance. Our results indicate that the larger 2048-dimension consistently outperformed the 512-dimension. These findings suggest that a larger feature channel dimension captures more nuanced information, leading to improved performance.

\subsubsection{Comprehensive ablation study of model components}

\begin{table}[t]
\centering
\label{tab:ablation_results}
\begin{tabular}{C{0.8cm}C{0.8cm}C{0.8cm}C{1.8cm}C{1.8cm}}
\toprule
\multicolumn{3}{c}{Method} & \multicolumn{1}{c}{1-shot} & \multicolumn{1}{c}{5-shot} \\
\cmidrule{4-5}
SCA & ICD & SCP &mAP@0.5 & mAP@0.5 \\
\midrule
 &  &  &60.7 & 62.8 \\
  &  & \ding{51} &65.3 & 67.2  \\
 & \ding{51} &  \ding{51} &46.2 & 68.3  \\
\ding{51}  &  & \ding{51} &45.9 & 68.1  \\
\ding{51}  & \ding{51} &  &64.2 & 64.8 \\
\ding{51}  & \ding{51} & \ding{51} & \textbf{68.4} & \textbf{70.2}  \\
\bottomrule
\end{tabular}
\caption{Ablation analysis of SCA, ICD, and SCP}
\label{tab:sis}
\end{table}

\begin{table}[t]
\centering
\begin{tabular}{C{2cm}C{2cm}C{2cm}}
\toprule
\multirow{2}{*}{Method} & \multicolumn{1}{c}{1-shot} & \multicolumn{1}{c}{5-shot} \\
\cmidrule{2-3}
 & mAP@0.5 & mAP@0.5 \\
\midrule
w/o smooth & 63.4 & 64.2 \\
w/o noise & 64.3 & 65.6 \\
fixed sigma & 58.3 & 60.8  \\
\textbf{Full} & \textbf{68.4} & \textbf{70.2} \\
\bottomrule
\end{tabular}
\caption{Ablation of label generator}
\label{tab:gpg}
\end{table}

\begin{figure}[t]
    \centering
    \includegraphics[width = \linewidth]{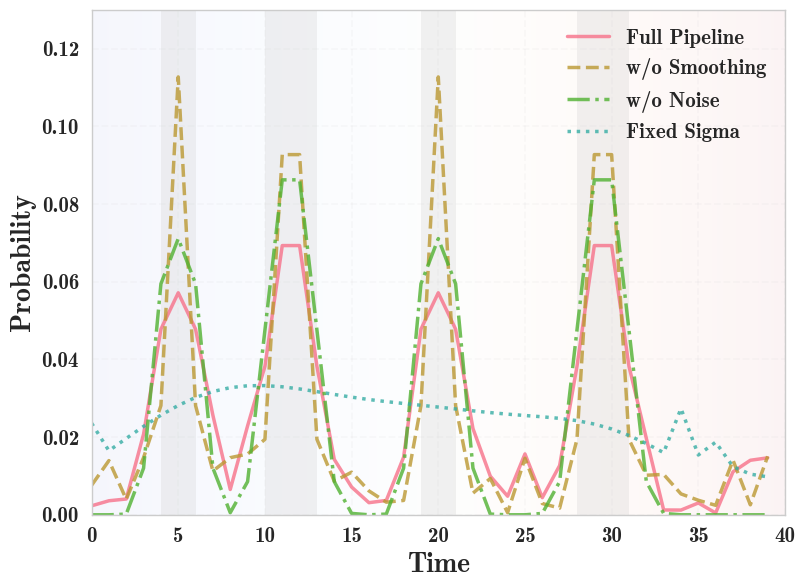}
    \caption{Ablation of label generator}
    \label{fig:mgpd}
\end{figure}

\begin{table}[t]
\centering
\label{tab:ablation_loss_weights}
\begin{tabular}{C{1cm}C{1cm}C{1cm}C{1.8cm}C{1.8cm}}
\toprule
\multicolumn{3}{c}{Loss} & \multicolumn{1}{c}{1-shot} & \multicolumn{1}{c}{5-shot} \\
\cmidrule{4-5}
$\boldsymbol{\mathcal{L}_{l1}}$ & $\boldsymbol{\mathcal{L}_{kl}}$ & $\boldsymbol{\mathcal{L}_{cls}}$ & mAP@0.5 & mAP@0.5 \\
\midrule
\ding{51} &  &  & 65.3 & 66.2 \\
 & \ding{51} &  & 62.9 & 64.3 \\
\ding{51} & \ding{51} &  & 67.6 & 68.7  \\
 & \ding{51} & \ding{51} & 66.9 & 68.1 \\
\ding{51} &  & \ding{51} & 63.6 & 64.6 \\
\ding{51} & \ding{51} & \ding{51} & \textbf{68.4} & \textbf{70.2} \\
\bottomrule
\end{tabular}
\caption{Ablation study: impact of loss}
\label{tab:iol}
\end{table}

Our model incorporates three key components: Spatial Contextual Aggregation (SCA), Inter-Channel Dependency (ICD), and Selective Cosine Penalization (SCP). To evaluate their individual and combined effects, we conducted a comprehensive ablation study, with results presented in Table \ref{tab:sis} for both 1-shot and 5-shot learning scenarios.

As is evident from the results, each component gives a significant contribution to the model's performance. SCA enhances spatial context understanding, ICD improves feature representation through channel dependencies while SCP refines temporal localization precision. The full model configuration (SCA + ICD + SCP) consistently achieves the highest performance.

\subsubsection{Ablation of label generator}

Table \ref{tab:gpg} presents an ablation study of our label generator algorithm. We evaluated three key components: smoothing operation, noise addition, and adaptive sigma calculation. The results demonstrate that each component contributes to the model's performance.

The smoothing operation proved crucial for generating coherent probability distributions across the temporal dimension while noise addition helped prevent overly confident predictions in low-probability regions. The adaptive sigma calculation outperformed a fixed sigma approach, highlighting its importance in handling actions of various durations. Figure \ref{fig:mgpd} visually illustrates the contribution of each component, providing an intuitive understanding of their effects on the generated labels. 

\subsubsection{Ablation of loss function}
We also discuss the important role of different parts in the loss. Table \ref{tab:iol} shows the $\mathcal{L}_{kl} \text{ and } \mathcal{L}_{l1}$ are crucial to the localization task because the performance of our model drop significantly without one of the $\mathcal{L}_{kl} \text{ and } \mathcal{L}_{l1}$. But only decrease slightly without the $\mathcal{L}_{cls}$ and the best result is under the combination of three losses.

This ablation study demonstrates the nuanced impact of loss weighting on the model's ability to accurately localize action boundaries and classify actions, particularly in few-shot learning contexts. 

\section{Conclusion}

In this work, we propose a novel method for few-shot multiple instances temporal action localization, which includes spatial-channel relation transformer, probability distribution learning, and interval clustering refinement. Our approach can accurately identify action boundaries with minimal labeled data, effectively capturing temporal and spatial contexts. The selective cosine penalty algorithm suppresses irrelevant boundaries while the probability learning and label generation enhance the model's ability to manage action duration diversity. Interval clustering ensures precise results, demonstrating our method's effectiveness in complex scenarios. We conduct comprehensive experiments on the benchmark ActivityNet1.3 and THUMOS14, demonstrating the competitiveness of our model's performance.

\bibliography{aaai25}

\begin{thebibliography}{55}
\providecommand{\natexlab}[1]{#1}

\bibitem[{Baum and Wilczek(1987)}]{baum1987supervised}
Baum, E.; and Wilczek, F. 1987.
\newblock Supervised learning of probability distributions by neural networks.
\newblock In \emph{Neural information processing systems}.

\bibitem[{Caba~Heilbron et~al.(2015)Caba~Heilbron, Escorcia, Ghanem, and Carlos~Niebles}]{caba2015activitynet}
Caba~Heilbron, F.; Escorcia, V.; Ghanem, B.; and Carlos~Niebles, J. 2015.
\newblock Activitynet: A large-scale video benchmark for human activity understanding.
\newblock In \emph{Proceedings of the ieee conference on computer vision and pattern recognition}, 961--970.

\bibitem[{Chao et~al.(2018)Chao, Vijayanarasimhan, Seybold, Ross, Deng, and Sukthankar}]{chao_rethinking_2018}
Chao, Y.-W.; Vijayanarasimhan, S.; Seybold, B.; Ross, D.~A.; Deng, J.; and Sukthankar, R. 2018.
\newblock Rethinking the {Faster} {R}-{CNN} {Architecture} for {Temporal} {Action} {Localization}.
\newblock In \emph{2018 {IEEE}/{CVF} {Conference} on {Computer} {Vision} and {Pattern} {Recognition}}, 1130--1139.

\bibitem[{Chen et~al.(2020)Chen, Gan, Shen, Huang, Zeng, and Tan}]{chen_relation_2020}
Chen, P.; Gan, C.; Shen, G.; Huang, W.; Zeng, R.; and Tan, M. 2020.
\newblock Relation {Attention} for {Temporal} {Action} {Localization}.
\newblock \emph{IEEE Transactions on Multimedia}, 22(10): 2723--2733.

\bibitem[{Dai et~al.(2017)Dai, Singh, Zhang, Davis, and Qiu~Chen}]{dai2017temporal}
Dai, X.; Singh, B.; Zhang, G.; Davis, L.~S.; and Qiu~Chen, Y. 2017.
\newblock Temporal context network for activity localization in videos.
\newblock In \emph{Proceedings of the IEEE International Conference on Computer Vision}, 5793--5802.

\bibitem[{Ester et~al.(1996)Ester, Kriegel, Sander, Xu et~al.}]{ester1996density}
Ester, M.; Kriegel, H.-P.; Sander, J.; Xu, X.; et~al. 1996.
\newblock A density-based algorithm for discovering clusters in large spatial databases with noise.
\newblock In \emph{kdd}, volume~96, 226--231.

\bibitem[{Feng et~al.(2018)Feng, Ma, Liu, Zhang, and Luo}]{feng2018video}
Feng, Y.; Ma, L.; Liu, W.; Zhang, T.; and Luo, J. 2018.
\newblock Video re-localization.
\newblock In \emph{Proceedings of the European Conference on Computer Vision (ECCV)}, 51--66.

\bibitem[{Gan and Zhang(2023)}]{Gan2023TemporalAP}
Gan, M.; and Zhang, Y. 2023.
\newblock Temporal Attention-Pyramid Pooling for Temporal Action Detection.
\newblock \emph{IEEE Transactions on Multimedia}, 25: 3799--3810.

\bibitem[{Gan, Zhang, and Su(2023)}]{Gan2023TemporalvisualPG}
Gan, M.; Zhang, Y.; and Su, S. 2023.
\newblock Temporal-visual proposal graph network for temporal action detection.
\newblock \emph{Applied Intelligence}, 53: 26008--26026.

\bibitem[{Gao, Chen, and Nevatia(2018)}]{gao2018ctap}
Gao, J.; Chen, K.; and Nevatia, R. 2018.
\newblock Ctap: Complementary temporal action proposal generation.
\newblock In \emph{Proceedings of the European conference on computer vision (ECCV)}, 68--83.

\bibitem[{Gao et~al.(2023)Gao, Cui, Zhao, Zhuo, Guan, and Wang}]{gao_novel_2023}
Gao, Z.; Cui, X.; Zhao, Y.; Zhuo, T.; Guan, W.; and Wang, M. 2023.
\newblock A {Novel} {Temporal} {Channel} {Enhancement} and {Contextual} {Excavation} {Network} for {Temporal} {Action} {Localization}.
\newblock In \emph{Proceedings of the 31st {ACM} {International} {Conference} on {Multimedia}}, 6724--6733. ACM.

\bibitem[{He, Li, and Lei(2021)}]{He2021FeaturePH}
He, J.; Li, G.; and Lei, J. 2021.
\newblock Feature Pyramid Hierarchies for Multi-scale Temporal Action Detection.
\newblock \emph{2020 25th International Conference on Pattern Recognition (ICPR)}, 2158--2165.

\bibitem[{He et~al.(2019)He, Zhu, Wang, Savvides, and Zhang}]{he2019bounding}
He, Y.; Zhu, C.; Wang, J.; Savvides, M.; and Zhang, X. 2019.
\newblock Bounding box regression with uncertainty for accurate object detection.
\newblock In \emph{Proceedings of the ieee/cvf conference on computer vision and pattern recognition}, 2888--2897.

\bibitem[{Hsieh et~al.(2023)Hsieh, Chen, Chang, and Liu}]{Hsieh_2023_WACV}
Hsieh, H.-Y.; Chen, D.-J.; Chang, C.-W.; and Liu, T.-L. 2023.
\newblock Aggregating Bilateral Attention for Few-Shot Instance Localization.
\newblock In \emph{Proceedings of the IEEE/CVF Winter Conference on Applications of Computer Vision (WACV)}, 6325--6334.

\bibitem[{Hu et~al.(2019)Hu, Mettes, Huang, and Snoek}]{hu2019silco}
Hu, T.; Mettes, P.; Huang, J.-H.; and Snoek, C.~G. 2019.
\newblock Silco: Show a few images, localize the common object.
\newblock In \emph{Proceedings of the IEEE/CVF International Conference on Computer Vision}, 5067--5076.

\bibitem[{Huang, Sugano, and Sato(2020)}]{huang2020improving}
Huang, Y.; Sugano, Y.; and Sato, Y. 2020.
\newblock Improving action segmentation via graph-based temporal reasoning.
\newblock In \emph{Proceedings of the IEEE/CVF conference on computer vision and pattern recognition}, 14024--14034.

\bibitem[{Jiang et~al.(2014)Jiang, Liu, Roshan~Zamir, Toderici, Laptev, Shah, and Sukthankar}]{THUMOS14}
Jiang, Y.-G.; Liu, J.; Roshan~Zamir, A.; Toderici, G.; Laptev, I.; Shah, M.; and Sukthankar, R. 2014.
\newblock {THUMOS} Challenge: Action Recognition with a Large Number of Classes.
\newblock \url{http://crcv.ucf.edu/THUMOS14/}.

\bibitem[{Lee, Jain, and Yun(2023)}]{Lee_2023_ICCV}
Lee, J.; Jain, M.; and Yun, S. 2023.
\newblock Few-Shot Common Action Localization via Cross-Attentional Fusion of Context and Temporal Dynamics.
\newblock In \emph{Proceedings of the IEEE/CVF International Conference on Computer Vision (ICCV)}, 10214--10223.

\bibitem[{Li et~al.(2022)Li, Zhang, Zhao, Feng, Yang, Liu, and Hou}]{Li2022PyramidRS}
Li, S.; Zhang, F.; Zhao, R.; Feng, R.; Yang, K.; Liu, L.-N.; and Hou, J. 2022.
\newblock Pyramid Region-based Slot Attention Network for Temporal Action Proposal Generation.
\newblock In \emph{British Machine Vision Conference}.

\bibitem[{Lin et~al.(2018)Lin, Zhao, Su, Wang, and Yang}]{lin2018bsn}
Lin, T.; Zhao, X.; Su, H.; Wang, C.; and Yang, M. 2018.
\newblock Bsn: Boundary sensitive network for temporal action proposal generation.
\newblock In \emph{Proceedings of the European conference on computer vision (ECCV)}, 3--19.

\bibitem[{Lin et~al.(2017)Lin, Goyal, Girshick, He, and Doll{\'a}r}]{lin2017focal}
Lin, T.-Y.; Goyal, P.; Girshick, R.; He, K.; and Doll{\'a}r, P. 2017.
\newblock Focal loss for dense object detection.
\newblock In \emph{Proceedings of the IEEE international conference on computer vision}, 2980--2988.

\bibitem[{Liu et~al.(2022)Liu, Wang, Hu, Tang, Zhang, Bai, and Bai}]{liu_end--end_2022}
Liu, X.; Wang, Q.; Hu, Y.; Tang, X.; Zhang, S.; Bai, S.; and Bai, X. 2022.
\newblock End-to-{End} {Temporal} {Action} {Detection} {With} {Transformer}.
\newblock \emph{IEEE Transactions on Image Processing}, 31: 5427--5441.

\bibitem[{Liu et~al.(2018)Liu, Ma, Zhang, Liu, and Chang}]{Liu2018MultiGranularityGF}
Liu, Y.; Ma, L.; Zhang, Y.; Liu, W.; and Chang, S.-F. 2018.
\newblock Multi-Granularity Generator for Temporal Action Proposal.
\newblock \emph{2019 IEEE/CVF Conference on Computer Vision and Pattern Recognition (CVPR)}, 3599--3608.

\bibitem[{Nag et~al.(2023)Nag, Zhu, Song, and Xiang}]{nag2023post}
Nag, S.; Zhu, X.; Song, Y.-Z.; and Xiang, T. 2023.
\newblock Post-processing temporal action detection.
\newblock In \emph{Proceedings of the IEEE/CVF Conference on Computer Vision and Pattern Recognition}, 18837--18845.

\bibitem[{Nag, Zhu, and Xiang(2021)}]{nag_few-shot_2021}
Nag, S.; Zhu, X.; and Xiang, T. 2021.
\newblock Few-{Shot} {Temporal} {Action} {Localization} with {Query} {Adaptive} {Transformer}.
\newblock ArXiv:2110.10552.

\bibitem[{Neubeck and Van~Gool(2006)}]{neubeck2006efficient}
Neubeck, A.; and Van~Gool, L. 2006.
\newblock Efficient non-maximum suppression.
\newblock In \emph{18th international conference on pattern recognition (ICPR'06)}, volume~3, 850--855.

\bibitem[{Perrett et~al.(2021)Perrett, Masullo, Burghardt, Mirmehdi, and Damen}]{perrett2021temporal}
Perrett, T.; Masullo, A.; Burghardt, T.; Mirmehdi, M.; and Damen, D. 2021.
\newblock Temporal-relational crosstransformers for few-shot action recognition.
\newblock In \emph{Proceedings of the IEEE/CVF conference on computer vision and pattern recognition}, 475--484.

\bibitem[{Rezatofighi et~al.(2019)Rezatofighi, Tsoi, Gwak, Sadeghian, Reid, and Savarese}]{Rezatofighi_2019_CVPR}
Rezatofighi, H.; Tsoi, N.; Gwak, J.; Sadeghian, A.; Reid, I.; and Savarese, S. 2019.
\newblock Generalized Intersection Over Union: A Metric and a Loss for Bounding Box Regression.
\newblock In \emph{Proceedings of the IEEE/CVF Conference on Computer Vision and Pattern Recognition (CVPR)}.

\bibitem[{Shou, Wang, and Chang(2016)}]{shou_temporal_2016}
Shou, Z.; Wang, D.; and Chang, S.-F. 2016.
\newblock Temporal {Action} {Localization} in {Untrimmed} {Videos} via {Multi}-stage {CNNs}.
\newblock In \emph{2016 {IEEE} {Conference} on {Computer} {Vision} and {Pattern} {Recognition} ({CVPR})}, 1049--1058.

\bibitem[{Snell, Swersky, and Zemel(2017)}]{snell2017prototypical}
Snell, J.; Swersky, K.; and Zemel, R. 2017.
\newblock Prototypical networks for few-shot learning.
\newblock \emph{Advances in neural information processing systems}, 30.

\bibitem[{Su, Wang, and Wang(2023)}]{Su2023MultiLevelCB}
Su, T.; Wang, H.; and Wang, L. 2023.
\newblock Multi-Level Content-Aware Boundary Detection for Temporal Action Proposal Generation.
\newblock \emph{IEEE Transactions on Image Processing}, 32: 6090--6101.

\bibitem[{Sung et~al.(2018)Sung, Yang, Zhang, Xiang, Torr, and Hospedales}]{sung2018learning}
Sung, F.; Yang, Y.; Zhang, L.; Xiang, T.; Torr, P.~H.; and Hospedales, T.~M. 2018.
\newblock Learning to compare: Relation network for few-shot learning.
\newblock In \emph{Proceedings of the IEEE conference on computer vision and pattern recognition}, 1199--1208.

\bibitem[{Tan et~al.(2021)Tan, Tang, Wang, and Wu}]{Tan2021RelaxedTD}
Tan, J.; Tang, J.; Wang, L.; and Wu, G. 2021.
\newblock Relaxed Transformer Decoders for Direct Action Proposal Generation.
\newblock \emph{2021 IEEE/CVF International Conference on Computer Vision (ICCV)}, 13506--13515.

\bibitem[{Tang et~al.(2023)Tang, Fan, Luo, Zhang, Zhang, and Yang}]{tang2023ddg}
Tang, X.; Fan, J.; Luo, C.; Zhang, Z.; Zhang, M.; and Yang, Z. 2023.
\newblock DDG-Net: Discriminability-Driven Graph Network for Weakly-supervised Temporal Action Localization.
\newblock In \emph{Proceedings of the IEEE/CVF International Conference on Computer Vision}, 6622--6632.

\bibitem[{Thatipelli et~al.(2022)Thatipelli, Narayan, Khan, Anwer, Khan, and Ghanem}]{thatipelli2022spatio}
Thatipelli, A.; Narayan, S.; Khan, S.; Anwer, R.~M.; Khan, F.~S.; and Ghanem, B. 2022.
\newblock Spatio-temporal relation modeling for few-shot action recognition.
\newblock In \emph{Proceedings of the IEEE/CVF Conference on Computer Vision and Pattern Recognition}, 19958--19967.

\bibitem[{Tran et~al.(2015)Tran, Bourdev, Fergus, Torresani, and Paluri}]{tran2015learning}
Tran, D.; Bourdev, L.; Fergus, R.; Torresani, L.; and Paluri, M. 2015.
\newblock Learning spatiotemporal features with 3d convolutional networks.
\newblock In \emph{Proceedings of the IEEE international conference on computer vision}, 4489--4497.

\bibitem[{Vinyals et~al.(2016)Vinyals, Blundell, Lillicrap, Wierstra et~al.}]{vinyals2016matching}
Vinyals, O.; Blundell, C.; Lillicrap, T.; Wierstra, D.; et~al. 2016.
\newblock Matching networks for one shot learning.
\newblock \emph{Advances in neural information processing systems}, 29.

\bibitem[{Wang et~al.(2021)Wang, Qing, Huang, Feng, Zhang, Jiang, Tang, Gao, and Sang}]{wang_proposal_2021}
Wang, X.; Qing, Z.; Huang, Z.; Feng, Y.; Zhang, S.; Jiang, J.; Tang, M.; Gao, C.; and Sang, N. 2021.
\newblock Proposal {Relation} {Network} for {Temporal} {Action} {Detection}.
\newblock ArXiv:2106.11812.

\bibitem[{Xia et~al.(2023)Xia, Wang, Shen, Zhou, Hua, and Tang}]{Xia2023ExploringAC}
Xia, K.; Wang, L.; Shen, Y.; Zhou, S.; Hua, G.; and Tang, W. 2023.
\newblock Exploring Action Centers for Temporal Action Localization.
\newblock \emph{IEEE Transactions on Multimedia}, 25: 9425--9436.

\bibitem[{Xu, Das, and Saenko(2017)}]{Xu2017RC3DRC}
Xu, H.; Das, A.; and Saenko, K. 2017.
\newblock R-C3D: Region Convolutional 3D Network for Temporal Activity Detection.
\newblock \emph{2017 IEEE International Conference on Computer Vision (ICCV)}, 5794--5803.

\bibitem[{Xu, Das, and Saenko(2019)}]{xu_two-stream_2019}
Xu, H.; Das, A.; and Saenko, K. 2019.
\newblock Two-{Stream} {Region} {Convolutional} {3D} {Network} for {Temporal} {Activity} {Detection}.
\newblock \emph{IEEE Transactions on Pattern Analysis and Machine Intelligence}, 41(10): 2319--2332.

\bibitem[{Yadan(2019)}]{Yadan2019Hydra}
Yadan, O. 2019.
\newblock Hydra - A framework for elegantly configuring complex applications.
\newblock Github.

\bibitem[{Yang, He, and Porikli(2018)}]{Yang2018OneShotAL}
Yang, H.; He, X.; and Porikli, F.~M. 2018.
\newblock One-Shot Action Localization by Learning Sequence Matching Network.
\newblock \emph{2018 IEEE/CVF Conference on Computer Vision and Pattern Recognition}, 1450--1459.

\bibitem[{Yang et~al.(2021)Yang, Wu, Wang, Jin, Xia, Yao, and Huang}]{Yang2021TemporalAP}
Yang, H.; Wu, W.; Wang, L.; Jin, S.; Xia, B.; Yao, H.; and Huang, H. 2021.
\newblock Temporal Action Proposal Generation with Background Constraint.
\newblock In \emph{AAAI Conference on Artificial Intelligence}.

\bibitem[{Yang et~al.(2020{\natexlab{a}})Yang, Peng, Zhang, Fu, and Han}]{yang_revisiting_2020}
Yang, L.; Peng, H.; Zhang, D.; Fu, J.; and Han, J. 2020{\natexlab{a}}.
\newblock Revisiting {Anchor} {Mechanisms} for {Temporal} {Action} {Localization}.
\newblock \emph{IEEE Transactions on Image Processing}, 29: 8535--8548.

\bibitem[{Yang et~al.(2020{\natexlab{b}})Yang, Hu, Mettes, and Snoek}]{vedaldi_localizing_2020}
Yang, P.; Hu, V.~T.; Mettes, P.; and Snoek, C. G.~M. 2020{\natexlab{b}}.
\newblock Localizing the {Common} {Action} {Among} a {Few} {Videos}.
\newblock In Vedaldi, A.; Bischof, H.; Brox, T.; and Frahm, J.-M., eds., \emph{Computer {Vision} – {ECCV} 2020}, volume 12352, 505--521.

\bibitem[{Yang, Mettes, and Snoek(2021)}]{yang_few-shot_2021}
Yang, P.; Mettes, P.; and Snoek, C. G.~M. 2021.
\newblock Few-{Shot} {Transformation} of {Common} {Actions} into {Time} and {Space}.
\newblock In \emph{2021 {IEEE}/{CVF} {Conference} on {Computer} {Vision} and {Pattern} {Recognition} ({CVPR})}, 16026--16035.

\bibitem[{Yin and Xiang(2023)}]{yin_enhanced_2023}
Yin, H.; and Xiang, X. 2023.
\newblock Enhanced {Multi}-scale {Transformer} {Network} for {Temporal} {Action} {Localization}.
\newblock In \emph{2023 {IEEE} {International} {Conference} on {Mechatronics} and {Automation} ({ICMA})}, 612--617.

\bibitem[{Yin et~al.(2023)Yin, Huang, Furuta, and Sato}]{Yin2023ProposalbasedTA}
Yin, Y.; Huang, Y.; Furuta, R.; and Sato, Y. 2023.
\newblock Proposal-based Temporal Action Localization with Point-level Supervision.
\newblock In \emph{British Machine Vision Conference}.

\bibitem[{Yuan et~al.(2016)Yuan, Ni, Yang, and Kassim}]{Yuan2016TemporalAL}
Yuan, J.-L.; Ni, B.; Yang, X.; and Kassim, A.~A. 2016.
\newblock Temporal Action Localization with Pyramid of Score Distribution Features.
\newblock \emph{2016 IEEE Conference on Computer Vision and Pattern Recognition (CVPR)}, 3093--3102.

\bibitem[{Zeng et~al.(2019)Zeng, Huang, Gan, Tan, Rong, Zhao, and Huang}]{zeng_graph_2019}
Zeng, R.; Huang, W.; Gan, C.; Tan, M.; Rong, Y.; Zhao, P.; and Huang, J. 2019.
\newblock Graph {Convolutional} {Networks} for {Temporal} {Action} {Localization}.
\newblock In \emph{2019 {IEEE}/{CVF} {International} {Conference} on {Computer} {Vision} ({ICCV})}, 7093--7102.

\bibitem[{Zeng et~al.(2022)Zeng, Huang, Tan, Rong, Zhao, Huang, and Gan}]{zeng_graph_2022}
Zeng, R.; Huang, W.; Tan, M.; Rong, Y.; Zhao, P.; Huang, J.; and Gan, C. 2022.
\newblock Graph {Convolutional} {Module} for {Temporal} {Action} {Localization} in {Videos}.
\newblock \emph{IEEE Transactions on Pattern Analysis and Machine Intelligence}, 44(10): 6209--6223.

\bibitem[{Zhang, Wu, and Li(2022)}]{zhang2022actionformer}
Zhang, C.-L.; Wu, J.; and Li, Y. 2022.
\newblock Actionformer: Localizing moments of actions with transformers.
\newblock In \emph{European Conference on Computer Vision}, 492--510.

\bibitem[{Zhang et~al.(2020)Zhang, He, Tu, Zhang, Han, and Yang}]{zhang_learning_2020}
Zhang, D.; He, L.; Tu, Z.; Zhang, S.; Han, F.; and Yang, B. 2020.
\newblock Learning motion representation for real-time spatio-temporal action localization.
\newblock \emph{Pattern Recognition}, 103: 107312.

\bibitem[{Zhao et~al.(2020)Zhao, Xiong, Wang, Wu, Tang, and Lin}]{zhao_temporal_2020}
Zhao, Y.; Xiong, Y.; Wang, L.; Wu, Z.; Tang, X.; and Lin, D. 2020.
\newblock Temporal {Action} {Detection} with {Structured} {Segment} {Networks}.
\newblock \emph{International Journal of Computer Vision}, 128(1): 74--95.

\end{thebibliography}

\end{document}